\documentclass[acmtog,authorversion]{acmart}

\usepackage[justification=justified,singlelinecheck=false]{subfig}
\usepackage{color}
\usepackage{hyperref}
\usepackage{tabularx}
\usepackage[scaled]{beramono}
\usepackage{fancyref}
\usepackage[T1]{fontenc}
\definecolor{kfcolor}{rgb}{0.75,0.0,0.0}
\definecolor{alexcolor}{rgb}{0.2,0.3,1.0}
\definecolor{willcolor}{rgb}{0.8,0.5,0.0}
\definecolor{mygray}{gray}{0.6}
\definecolor{keywordcolor}{rgb}{0.0,0.0,1.0}
\definecolor{typecolor}{rgb}{0.17,0.56,0.68}
\definecolor{commentcolor}{rgb}{0.0,0.5,0.0}
\definecolor{ratecolor}{gray}{0.4}

\newcommand{\scanner}{Scanner}

\newcommand{\smallcaps}[1]{\textsc{#1}}
\newcommand{\hist}{\smallcaps{hist}}
\newcommand{\dnn}{\smallcaps{dnn}}
\newcommand{\flow}{\smallcaps{flow}}
\newcommand{\pose}{\smallcaps{pose}}

\newcommand{\imgcpu}{\smallcaps{imgcpu}}
\newcommand{\vidcpu}{\smallcaps{vidcpu}}
\newcommand{\vidgpu}{\smallcaps{vidgpu}}

\newcommand{\decvideo}{\smallcaps{base}}
\newcommand{\decsmallgop}{\smallcaps{smallgop}}
\newcommand{\decstrided}{\smallcaps{strided}}

\newcommand{\decode}{\smallcaps{stride-1}}
\newcommand{\stride}{\smallcaps{stride-24}}
\newcommand{\gath}{\smallcaps{gather}}
\newcommand{\range}{\smallcaps{range}}
\newcommand{\keyframe}{\smallcaps{keyframe}}

\newcommand{\cinema}{\smallcaps{cinema}}
\newcommand{\tvnews}{\smallcaps{tvnews}}

\usepackage{trivfloat}
\trivfloat{listing}

\usepackage{listings}
\lstdefinelanguage{pseudo}{%
	morekeywords=[0]{def,for,each,for,if,break,continue,return,loop,from,to,by,in,select,from,where},
	morecomment=[l]{\#}
}

\lstdefinelanguage{GLSL}{%
	morekeywords=[0]{if,void,else,uniform,varying,for,in,out,shader,using,pipeline,group,abstract,finalizes,optional,final,return,world,export,import,alt},
	morekeywords=[1]{float,vec2,vec3,vec4,mat2,mat3,mat4,int,sampler2D},
	morekeywords=[2]{@MeshVertex,@MaterialUniform,@Vertex,@Fragment,@FrameUniform,@PrebakeTex,@PrebakeTexVs,@Frame,@Eye,@PrebakeVertex,@PrebakeUniform},
	morecomment=[l]{//},
}

\newcommand{\code}[1]{\text{\lstinline[basicstyle=\ttfamily]{#1}}}

\usepackage{booktabs} 

\acmPrice{15.00}

\setcopyright{acmlicensed}
\acmJournal{TOG}
\acmYear{2018}
\acmVolume{37}
\acmNumber{4}
\acmArticle{138}
\acmMonth{8}
\acmDOI{10.1145/3197517.3201394}
\acmSubmissionID{577}

\setcitestyle{square}

\begin{document}

\title{Scanner: Efficient Video Analysis at Scale}

\author{Alex Poms}
\affiliation{\institution{Carnegie Mellon University}\country{USA}}
\author{Will Crichton}
\author{Pat Hanrahan}
\author{Kayvon Fatahalian}
\affiliation{\institution{Stanford University}\country{USA}}

\renewcommand{\shortauthors}{Poms, Crichton, Hanrahan, and Fatahalian}

\begin{abstract}
A growing number of visual computing applications depend on the analysis of large video collections.
The challenge is that scaling applications to operate on these datasets requires
efficient systems for pixel data access and parallel processing across large numbers of machines.
Few programmers have the capability to operate efficiently at these scales,
limiting the field's ability to explore new applications that leverage big video data.
In response, we have created \scanner, a system for productive and efficient video analysis at scale.
\scanner\ organizes video collections as tables in a data store optimized for sampling frames from compressed video,
and executes pixel processing computations, expressed as dataflow graphs, on these frames.
\scanner\ schedules video analysis applications expressed using these abstractions onto heterogeneous 
throughput computing hardware, such as multi-core CPUs, GPUs, and media processing ASICs, for high-throughput pixel processing.
We demonstrate the productivity of \scanner\ by authoring a variety of video processing applications
including the synthesis of stereo VR video streams from multi-camera rigs,
markerless 3D human pose reconstruction from video,
and data-mining big video datasets such as hundreds of feature-length films or over 70,000 hours of TV news.
These applications achieve near-expert performance on a single machine and scale efficiently to hundreds of machines,
enabling formerly long-running big video data analysis tasks to be carried out in minutes to hours.

\end{abstract}




\maketitle

\section{Introduction}

The world is increasingly instrumented with sources of video:
cameras are commonplace on people (smartphone cameras, GoPros), on
vehicles (automotive cameras, drone videography), and in urban
environments (traffic cameras, security cameras). Extracting
value from these high-resolution video streams is a key research and
commercial challenge, and a growing number of applications in fields
like computer graphics, vision, robotics and basic science
are based on analyzing large amounts of video.

The challenge is that scaling video analysis tasks to large video
collections (thousands of hours of cable TV or YouTube clips,
the output of a modern VR video capture rig)
requires optimized systems for managing pixel data as well as
efficient, parallel processing on accelerated computing hardware
(clusters of multi-core CPUs, GPUs, and ASICs).
Unfortunately, very few programmers have the skill set to
implement efficient software for processing large video datasets, inhibiting the field's ability to explore new applications that leverage this data.
Inspired by the impact of data analytics frameworks such as
MapReduce\,\cite{Dean:2004:MapReduce} and
Spark\,\cite{Zaharia:2010:Spark}, which facilitate rapid development of scalable big-data analytics applications,
we have created \emph{\scanner}, a system for productive and efficient \emph{big video data} analysis.

\scanner\ provides integrated system support for two performance-critical aspects of video analysis: storing and accessing pixel data from large video collections, and executing expensive pixel-level operations in parallel on large numbers of video frames.
\scanner\ addresses the first need by organizing video collections and derived raster data (depth maps, activation maps, flow fields, etc.) as tables in a data store whose implementation is optimized for compressed video.  It addresses the second need by organizing pixel-analysis tasks as dataflow graphs that operate on sequences of frames sampled from tables. \scanner\ graphs support features useful for video processing, such as sparse sampling of video frames, access to temporal windows of frames, and state propagation across computations on successive frames. \scanner\ schedules these computations efficiently onto  heterogeneous computing hardware such as multi-core CPUs, GPUs, and media processing ASICs.

We demonstrate that applications using \scanner\ for expensive, pixel-level video processing operations achieve near-expert performance when deployed on workstations with high-core count CPUs and multiple GPUs. The same applications also scale efficiently to hundreds of machines without source-level change. We report on experiences using \scanner\ to implement several large-scale video analysis applications including VR video processing, 3D human pose reconstruction from multi-viewpoint video, and data mining large video datasets of TV news. In these cases, \scanner\ enabled video analysis tasks that previously required days of processing (when implemented by researchers and data scientists using ad hoc solutions) to be carried out efficiently in hours to minutes. Scanner is available as open-source code at \href{https://github.com/scanner-research/scanner}{https://github.com/scanner-research/scanner}.

\section{Challenges of Video Analysis}
\label{sec:goals}

\begin{figure*}[t!]
  \centering
  \includegraphics[width=7in]{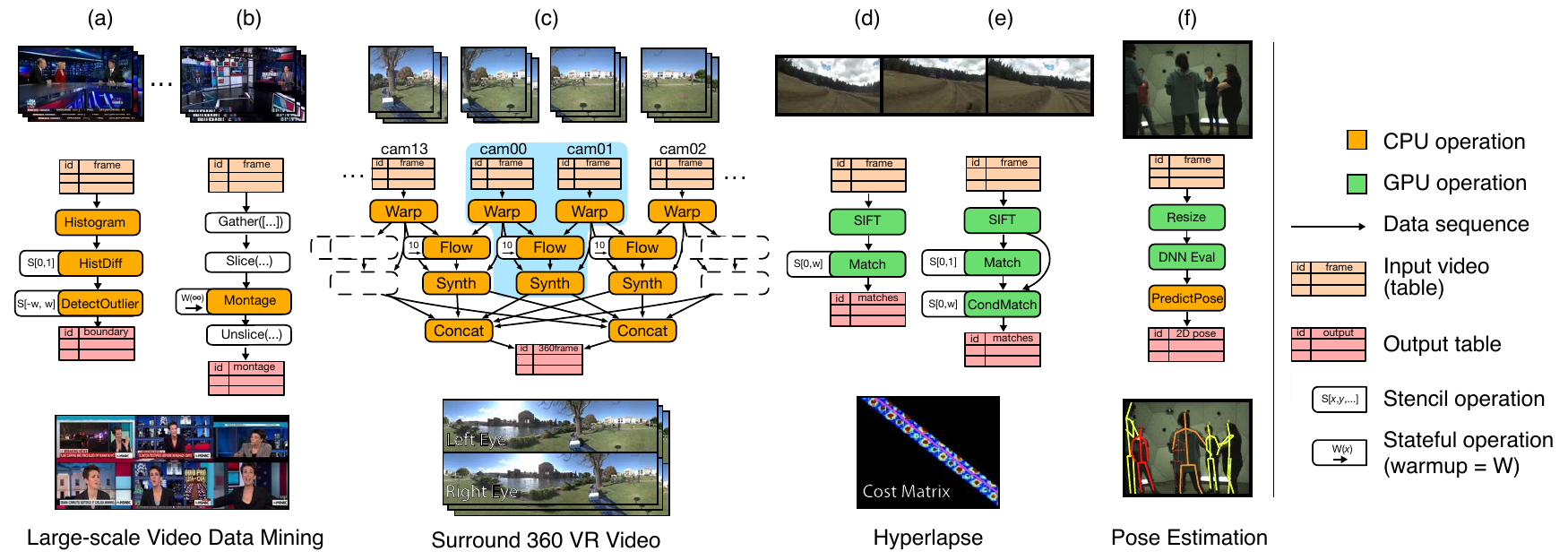}
  \vspace{-1.0em}
  \caption{We have implemented a set of video analysis applications in \scanner\ by expressing key pixel processing operations as dataflow graphs (Section~\ref{sec:computationgraph}).  Each application contributes unique challenges to \scanner's design, such as stateful processing, combining information across video streams, sparse frame access, and the need to process large numbers of video clips. Image credit, left to right: \textit{The Rachel Maddow Show} \textcopyright~MSNBC 2015-2017, "Palace of Fine Arts Take 1" \textcopyright~Facebook 2017, ``Run 5K'' clip (top) and Figure 1 (bottom) from \cite{Joshi:2015:hyperlapse}, ``160422\_mafia2'' scene from \cite{Joo:2016:Panoptic}.}
  \label{fig:appsgraphs}
  \vspace{-1.0em}
\end{figure*}

Executing pixel-analysis pipelines (e.g., feature extraction, face/object detection, image similarity and alignment) on large image collections is the performance-critical component of many big visual data applications
such as data-driven image manipulation
and enhancement\,\cite{Hays:2007,Kemelmacher-Shlizerman:2016:transfiguring},
novel techniques for organizing and browsing photo
collections\,\cite{Snavely:2006:phototourism,Sivic:2008:inifinite},
and exploratory data mining of the visual world\,\cite{Doersch:2012:Paris,Chen:2013:Neil,Zhu:2014:AverageExplorer,Ginosar:2017:portraits,Matzen:2017:streetstyle}.
While these early applications analyzed collections of images, a growing class of applications now seek to manipulate large video datasets. To better understand the challenges and requirements of these video analysis workloads, we selected a diverse set of video analysis applications to guide the design of \scanner.  Fig.~\ref{fig:appsgraphs} summarizes the structure of these applications, which are implemented in \scanner\ and evaluated at scale in Section~\ref{sec:applications}.

\subsection{Workloads}
\label{sec:workloads}

\textbf{Large-scale video data mining.}  Many applications now seek to perform labeling and data-mining of large video collections. Examples include autonomous vehicle development\,\cite{Bojarski:2016:selfdriving}, surveillance, smart-city monitoring, and everyday egocentric video capture\,\cite{Singh:2016:Kcam}. These computations require both traditional computer vision operations (optical flow, object tracking, etc.) and DNN inference (object detection, frame segmentation, activity recognition) to be executed on millions to billions of video frames.  To keep costs manageable, it is common to sparsely sample frames from the video (e.g. every $n$-th frame, a list of frames likely to contain interesting objects). In Section~\ref{sec:analyticsapps} we report on experiences labeling and data mining two large video datasets: a dataset containing over 600~feature length films (106~million frames) and a dataset of 70,000 hours of TV news (12 billion frames, 20 TB).

\textbf{360-degree stereo video generation for VR.} Software for generating omnidirectional stereo (ODS) video, 360 degree stereo panoramas, provide a solution for authoring VR video. We ported the Surround~360 pipeline\,\cite{Surround360} for producing ODS video video from 14 synchronized 2K video streams. This application involves \emph{per-frame operations} (warping input frames to a spherical projection), \emph{cross-video-stream operations} (depth estimation between frames from adjacent cameras), within-stream frame-to-frame dependencies (stateful temporal smoothing of computed flow fields), and the ability to output a final compressed high-resolution video stream. Surround~360 processing is computationally intense; it can take over twelve seconds to produce a single output frame on a 32-core server. The Jump VR Video processing pipeline has similar characteristics\,\cite{Anderson:2016:Jumpvr}.

\textbf{Hyperlapse generation.}  \textit{Hyperlapses} are stabilized timelapse videos synthesized from long videos captured with moving cameras. The challenge of generating a high-quality hyperlapse involves selecting source video frames that approximate a desired timelapse playback speed while minimizing apparent camera movement. We have implemented two variants of the frame-selection computation described by Joshi~et~al.\,\protect\shortcite{Joshi:2015:hyperlapse}, which performs SIFT feature extraction and matching over \emph{sliding windows} of frames from a video stream (temporal stencil computations).

\textbf{3D human pose estimation.}  Recent computer vision advances make it possible to estimate temporally consistent human joint locations from dense multi-viewpoint video. This offers the promise of markerless human motion capture, even in high-occlusion scenarios, but comes at the cost of processing many video streams. For example, human motion capture sessions from the CMU Panoptic Dataset\,\cite{Joo:2015:Panoptic} feature 480 synchronized streams of 640$\times$480 video (see visualization in Fig.~\ref{fig:appsviz}).  The dominant cost of a top-performing method for 3D pose reconstruction from these streams\,\cite{Joo:2016:Panoptic} involves evaluating a DNN on every frame of all streams to estimate 2D pose. The 2D poses are subsequently fused to obtain a 3D pose estimate.

\subsection{Challenges}
\label{sec:challenges}

\scanner's goal is to enable rapid development and scaling of applications such as those described above. This required a system with flexible abstractions to span a range of video analysis tasks, but also sufficiently constrained to allow efficient, highly-parallel implementations. Specifically, our experiences implementing the applications in Section~\ref{sec:workloads} suggest that the size and temporal nature of video introduces several unique system requirements and challenges:

\textbf{Organize and store compressed video.}
Managing tens of thousands of video clips, as well as per-frame raster products derived from its analysis (e.g., multiple resolutions of frames, flow fields, depth maps, feature maps, etc.), can be tedious and error prone without clear abstractions for organizing this data. The relational data model\,\cite{Codd:1970:relational} provides a natural representation for organizing video collections (e.g., a table per video, a row per video frame), however we are not aware of a modern database system optimized for managing, indexing, and \emph{providing efficient frame-level access} to data stored compactly using video-specific compression (e.g., H.264).  While some applications require video data to be maintained in a lossless form, in most cases it is not practical to store large video datasets as individual frames (even if frames are individually compressed). Video collections can fill TBs of storage even when encoded compactly using video-specific compression schemes. Ignoring inter-frame compression opportunities can increase storage footprint by an order or magnitude or more.

\textbf{Support a flexible set of frame access patterns.}
Video compression schemes are designed for sequential frame access during video playback, however video analysis tasks exhibit a rich set of frame access patterns. While some applications access all video frames, others sample frames sparsely, select frame ranges, operate on sliding windows (e.g., optical flow, stabilization), or require joining frames from multiple videos (e.g., multi-view stereo). A system for video analysis must provide a rich set of streaming frame-level access patterns and implement these patterns efficiently on compressed video representations.

\textbf{Support frame-to-frame (temporal) dependencies.}
Reasoning about a sequence of video frames as a whole (rather than considering individual frames in isolation) is fundamental to algorithms such as object tracking, optical flow, or activity recognition.  Sequence-level reasoning is also key to achieving greater algorithmic efficiency when executing per-frame computations on a video stream since it is possible to exploit frame-to-frame coherence to accelerate analysis.  Therefore, the system must permit video analysis computations to maintain state between processing of frames, but also constrain frame-to-frame dependencies to preserve opportunities for efficient data streaming and parallel execution.

\textbf{Schedule pixel-processing pipelines (with black-box kernels) onto heterogeneous, parallel hardware.}
Authoring high-performance implementations of low-level image processing kernels (e.g., DNN evaluation, feature extraction, optical flow, object tracking) is difficult, so application developers typically construct analysis pipelines from pre-existing kernels provided by state-of-the-art performance libraries (e.g., cuDNN, OpenCV) or synthesized by high-performance DSLs (e.g., Halide\,\cite{Ragan-Kelley:2012:Halide}).  Therefore, a video analysis system must assume responsibility for automatically scheduling these pipelines onto parallel, heterogeneous machines, and orchestrate efficient data movement between kernels. (The 3D human pose reconstruction pipeline presented in Section~\ref{sec:poseReconstruction} involves computation on the CPU, GPU, and video decoding ASICs.)
Although a single system for both kernel code generation and distributed execution provides opportunities for global optimization, it is not practical to force applications to use a specific kernel code generation framework. For reasons of productivity and performance, \scanner\ should minimally constrain what 3rd-party kernels applications can use.

\textbf{Scaling video analysis.} Designing abstractions to address the above challenges is difficult because they must also permit an implementation which is able to scale from a workstation packed with GPUs underneath a researcher's desk to a cluster of thousands of machines, and from a dataset of a few 4K video streams to millions of 480p videos. Specifically, our examples require \scanner\ to scale in a number of ways:

\begin{itemize}

\item \textbf{Number of videos.} \scanner\ applications should scale to video datasets of arbitrary size (in our cases: millions or billions of frames), and consisting of both long videos (many feature length films or long-running vehicle capture sessions), or a large number of short video clips (e.g., millions of YouTube video clips).

\item \textbf{Number of concurrent video streams.} We seek to handle applications that must process and combine a large number of video streams capturing a similar subject, scene, or event, such as VR video (14 streams) and 3D pose reconstruction (480 streams) discussed in Section~\ref{sec:workloads}. \scanner\ should accelerate computationally intensive pipelines to enable processing these streams at near-real time rates.

\item \textbf{Number of throughput-computing cores.} \scanner\ applications should efficiently utilize throughput computing hardware (multi-core CPUs, multiple GPUs, media processing ASICs, and future DNN accelerators\,\cite{jouppi:2017:tpu}) to achieve near-expert performance on a single machine, and also scale out to large numbers of compute-rich machines (thousands of CPUs or GPUs) with little-to-no source-level change.

\end{itemize}

\begin{figure*}[t!]
  \centering
  \includegraphics[width=7in]{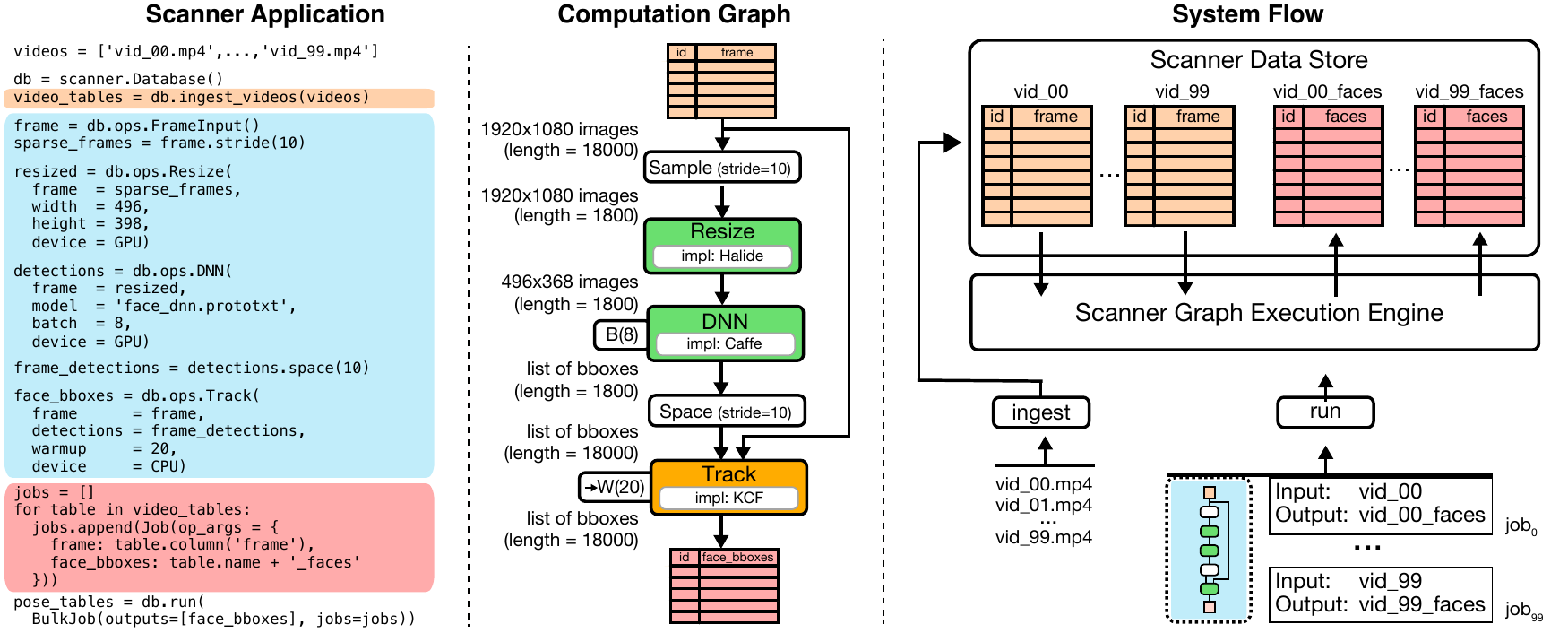}
  \vspace{-1.00em}
  \caption{\scanner\ computation graphs (blue) operate on sequences of per-frame data extracted from data store tables (tan), and produce outputs that are stored as new tables (pink).  This graph performs expensive face detection every 10th frame, and uses these detections to seed an object tracker run on each frame.}
  \label{fig:overview}
  \vspace{-1.25em}
\end{figure*}

We have designed \scanner\ to address these challenges.  When our goals of productivity, scope, and performance conflict, we opted in favor of maintaining a scalable and performant system. This philosophy resulted in a number of clear \emph{non-goals} for \scanner. For example, \scanner\ does not seek to aid with processing the results of pixel or feature-level analysis (image metadata, object labels, histograms, etc.). Post-processing these smaller derived data sets often involves a diverse set of algorithms that are well supported by existing data analysis frameworks.
Also, \scanner\ does not seek to define its own programming language for authoring high-performance kernel functions. Many domain-specific programming frameworks exist for this purpose today and \scanner\ aims to inter-operate with and augment these best-in-class tools, not replicate their functionality.

\section{\scanner\ Concepts}
\label{sec:concepts}

In this section we describe the primary abstractions used to construct \scanner\ applications. \scanner\ adopts two dataflow programming concepts familiar to users of existing data analytics frameworks and stream processing systems\,\cite{Dean:2004:MapReduce,Zaharia:2010:Spark,TensorFlow:2016,MxNet:2015}, but extends and implements these concepts uniquely for the needs of efficient video processing.

\textbf{Videos as logical tables.} \scanner\ represents video collections and the pixel-level products of video frame analysis (e.g., flow fields, depth maps, activations) as tables in a data store. \scanner's data store features first-class support for video frame column types to facilitate key performance optimizations.

\textbf{Video processing operations as dataflow graphs.} \scanner\ structures video analysis tasks as dataflow graphs whose nodes produce and consume sequences of per-frame data. \scanner's embodiment of the dataflow model includes operators useful for common video processing tasks such as sparse frame sampling, stenciled frame access, and stateful processing across frames.

We first provide an example of how \scanner's abstractions are used to conduct a simple video analysis task, then describe the motivation and design of key system primitives in further detail.

\subsection{Scanner Workflow}
\label{sec:workflow}

Fig.~\ref{fig:overview} illustrates a simple video analysis application (implemented using \scanner's Python API) that annotates a video with bounding boxes for the faces in each frame.

First, the application ingests a collection of videos into the \scanner\ data store, shown in yellow.  Logically, each video is represented by a table, with one row per video frame. In the example, ingest produces 100 tables, each with 18,000 rows, corresponding to 10-minute 30~FPS videos. The \scanner\ data store provides first-class support for table columns of \emph{video frame} type, which facilitates compact storage and efficient frame-level access to compressed video data (Section~\ref{sec:sparsedecode}).  (See supplemental material for additional detail on how first-class video support enables \scanner's storage formats to be optimized to specific access patterns without needing application-level change.)

Next, the application defines a five-stage \emph{computation graph} that specifies what processing to perform on the video frames (code shaded in blue). Since accurate face-detection is costly, the application samples every 10th frame from the input video (\code{Stride}), downsamples the resulting frames (\code{Resize}), then evaluates a DNN to detect faces in each downsampled frame to produce a per-frame list of bounding boxes (\code{DNN}).  The 3~FPS (sparse-in-time) detections are then re-aligned (\code{Space}) with the original high-resolution, 30~FPS image sequence from the data store, and used to seed an object tracker (\code{Track}) that augments the original detections with additional detections produced by tracking on the original frames.  The computation graph outputs a sequence of per-frame face bounding boxes that is stored as a new table with a column named \code{face_bboxes}.

A \scanner\ \emph{job} specifies a computation graph to execute and the tables it consumes and produces. In this example, the application defines one job for each video (code shaded in pink). \scanner\ automatically schedules all jobs onto a target machine (potentially exploiting parallelism across jobs, frames in a job, and computations in the graph), resulting in the creation of new database tables (shown to the right in pink in Fig.~\ref{fig:overview}).  After using \scanner\ to perform the expensive pixel processing operations on video frames, an application typically exports results from \scanner, and uses existing data analysis frameworks to perform less performance-critical post-processing of the face bounding box locations.

\subsection{Computation Graphs}
\label{sec:computationgraph}

\begin{figure*}[t!]
  \centering
  \includegraphics[width=7in]{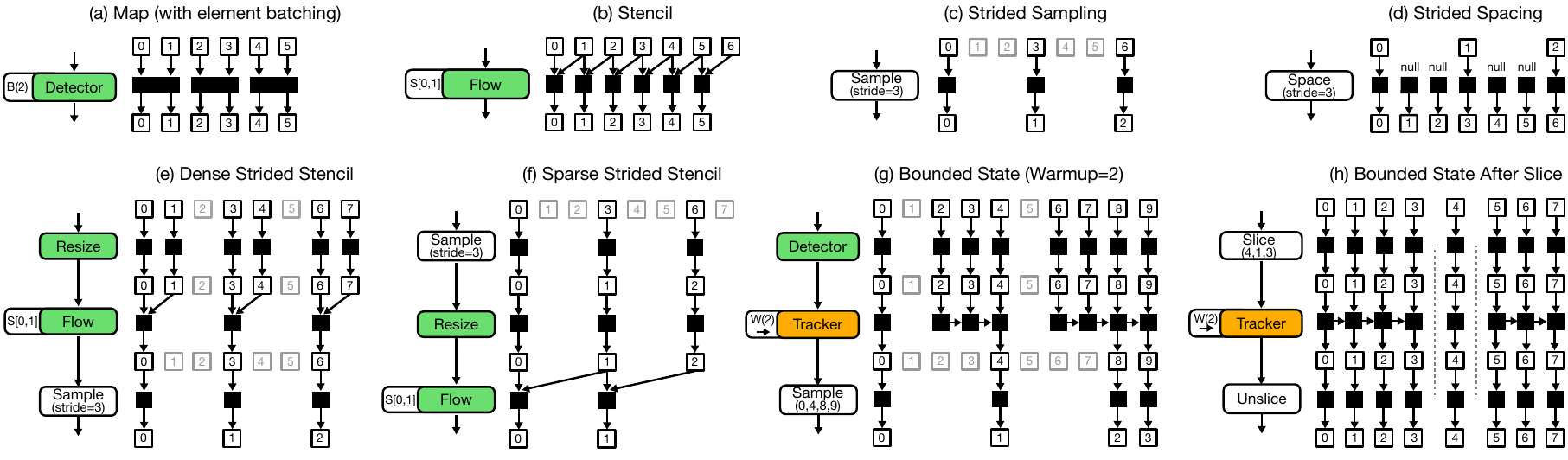}
  \vspace{-1.25em}
  \caption{\scanner\ analyzes operation dependencies to reduce computation during graph execution. White boxes denote elements of a sequence, and are labeled with their corresponding sequence domain point. Black boxes denote the execution of a graph operation on an element. Grayed elements are not required to produce the graph's required outputs and need not be computed.}
  \label{fig:sampling}
  \vspace{-1.25em}
\end{figure*}

\scanner\ applications express video processing tasks in the dataflow model by defining computation graphs. For consistency with\,\cite{TensorFlow:2016,MxNet:2015}, we refer to graph nodes, which define stages of computation, as \emph{operations}.  Graph edges are \emph{sequences} whose elements contain per-frame data communicated between operations.  Figures~\ref{fig:appsgraphs} and \ref{fig:overview} illustrate \scanner\ computation graphs for our example applications.  These graphs range from simple pipelines defining stages of processing on a single video to complex DAGs of many operations on multiple input video streams.

\textbf{Sequences.} \scanner\ sequences are finite-length, 1D collections that are
streamed element-by-element (or in small batches) to graph operations\,\cite{Buck:2004:Brook,Thies:2002:Streamit,Zaharia:2010:Spark}.  Each element in a length $N$ sequence is associated with a point in the $[0,N)$ domain.  It is typical for sequence elements in \scanner\ applications to be video frames, or derived structures produced by graph operations, such as transformed images, flow fields, depth maps, or frame metadata (e.g., lists of per-frame object bounding boxes).

\textbf{Graph Operations.} A major challenge in \scanner's design was selecting a set of graph operations that could be composed to express a rich set of video processing applications, but was sufficiently constrained to enable a streaming, data-parallel implementation.  \scanner\ supports the following classes of graph operations, which are characterized by their input stream access patterns, and whether state is propagated between invocations on consecutive stream elements.

\emph{Maps.}  \scanner\ operations may be mapped (Fig.~\ref{fig:sampling}-a) onto input sequences or onto multiple sequences of the same length (e.g., resizing an input frame or evaluating a DNN to generate per-frame activations).

\emph{Sampling/spacing operations.}  Sampling and spacing operations (Figure~\ref{fig:sampling}-c,d) modify the length of sequences by selecting a subset of elements from the input sequence (sampling) or adding ``fill'' elements to it (spacing). Sampling operations enable computation on a sparse set of frames for computational efficiency or when specific frames must be selected for processing. For example, sampling every 30th row from a table representing a one-minute long, 30~FPS video (1800 frames) yields a length~60 sequence representing the video sampled at one frame per second.  Spacing operations invert sampling and are used to align sequences representing data sampled at different frame rates. For example, in Fig.~\ref{fig:overview} a spacing operation was used to convert face detections computed at 3~FPS back into a 30~FPS stream. Both sampling and spacing operations can be defined by strides, ranges, or index lists.

\emph{Stencil operations.} Stencil operations gain access to a window of elements from the input sequence defined by a constant-offset stencil.  For example, the optical flow operation in Fig.~\ref{fig:sampling}-b  requires elements $i$ and $i+1$ of the input sequence to generate output element $i$ (the stencil is denoted by S[0,1] next to the operation).  Composing stencil and sampling operations yields a rich set of frame access patterns.  For example, performing stride-$N$ sampling prior to optical flow with stencil $(i,i+1)$ yields flow vectors computed on a low frame rate video sequence (Fig.~\ref{fig:sampling}-f), whereas sampling after the flow operation yields a sparse set of flow fields computed from differences between original video frames (Fig.~\ref{fig:sampling}-e).

\emph{Bounded State Operations.} Video processing requires operations that maintain state from frame-to-frame, either because it is fundamental to the operation being performed (e.g., tracking) or as a compute optimization when there is little frame-to-frame change.  However, if unconstrained, stateful processing would force serialization of graph execution.  As a compromise, \scanner\ allows stateful operations, but limits the extent to which the processing of one sequence element can affect processing of later ones.  Specifically, \scanner\ guarantees that prior to invoking an instance of a bounded state operation to generate output element $i$, the operation will have previously been invoked to produce \emph{at least} the previous $W$ elements of its output sequence.  (The ``warmup'' value $W$ is provided to \scanner\ by the stateful operation.)  As a result, the operation is guaranteed that effects of processing element $i$ will be visible when processing elements ($i+1$,...,$i+W-1$)  (Fig.~\ref{fig:sampling}-g: horizontal arrows).  In Figs. \ref{fig:appsgraphs} and \ref{fig:sampling}, we denote the warmup size of bounded state operations (in elements) using the notation \code{W()}. An operation may have an infinite warmup, indicating that it must process input sequences serially (zero parallelism).

Warmup allows operations to benefit from element-to-element state propagation, while the bound on information flow provides \scanner\ flexibility to parallelize stateful operators at the cost of a small amount of redundant computation.  For example, it is valid to execute a bounded state operation ($W=2$) with a length-100 output sequence by producing output elements [0,50) on one machine independently from elements [48,99) on a second.  \scanner\ automatically discards warmup elements 48 and 49 from the second worker (it does not include them in the output sequence), although effects of their processing may impact the value of subsequent elements (e.g., 50) generated by this worker. Bounded state operations use warmup to approximate the output of unbounded (fully serial) stateful execution when the influence of an operation's effects is known to be localized in the video stream.  For example, warmup of a few elements can be used to prime an object tracker prior to producing required outputs, or to minimize temporal discontinuities in the outputs of a stateful operation at the boundary of two independently computed regions.

    \emph{Slicing/unslicing operations.}  Slicing and unslicing operations insert and remove boundaries that affect stenciling and state propagation in a sequence.  For example, slicing a video sequence at intervals according to shot boundaries would reset stencil operation access patterns and stateful processing to avoid information flow between shots (Fig.~\ref{fig:sampling}-h illustrates the use of slicing to partition a sequence into three independent slices).  Unslicing removes these boundaries for all subsequent operations.  Slicing and unslicing can be viewed as a constrained form of sequence nesting.

 \textbf{Computation Graph Limitations.}  \scanner's design constrains the data flow expressible in computation graphs to permit two performance-critical graph scheduling optimizations: parallel graph execution and efficient graph scheduling in conditions of sparse sampling (Section~\ref{sec:impl}).  For similar reasons, \scanner\ currently disallows computation graphs with loops or operations that perform data-dependent filtering (discarding elements that do not pass a predicate) or amplification. Although \scanner\ operations are not provided mechanisms for dynamically modifying sequence length, sequence elements can be of tuple or list type (e.g., operations can produce variable length lists of face bounding boxes per frame).

 \subsection{Defining Graph Operations}
 \label{sec:operationapi}

Consistent with the goals from Section~\ref{sec:goals}, \scanner\ does not provide mechanisms for defining the implementation of graph operations. With the exception of system-provided sampling, spacing, and slicing operations, \scanner\ operation definitions are implemented in 3rd party languages, externally compiled, and exposed to applications as \scanner\ graph operations using an operation definition API inspired by that of modern dataflow frameworks for machine learning\,\cite{TensorFlow:2016,MxNet:2015}.  In the face detection example from Fig.~\ref{fig:overview}, \code{Resize} is implemented in Halide, \code{DNN} by the Caffe library\,\cite{Jia:2014:caffe} in CUDA, and \code{Track} as multi-threaded C++.  For bounded state operations, the allocation and management of mutable state carried across invocations is encapsulated entirely within the operation's definition and is opaque to \scanner\ (e.g., internal object tracker state).

Although \scanner\ is oblivious to the details of an operation's implementation, to facilitate efficient graph scheduling, all operations must declare their processing resource requirements (e.g., requires a GPU, requires $N$ CPU cores) and data dependencies (warmup amount for stateful operations, stencil offsets for stencil operations) to \scanner.  For efficiency, \scanner\ also supports operations that generate a batch of output elements (rather than a single element) per invocation (e.g., DNN inference on a batch of frames).  We denote the batch size of operations as \code{B()} in computation graph illustrations.

\section{Runtime Implementation}
\label{sec:impl}

\scanner\ jobs are executed by a high-performance runtime that provides applications high-throughput access to video frames and efficiently schedules computation graphs onto a parallel machine. While aspects of \scanner's implementation constitute intelligent application of parallel systems design principles, the challenges of efficiently accessing compressed video data and executing compositions of sampling, stenciling, and bounded state graph operations led to unique implementation choices detailed here.

\subsection{Graph Scheduling and Parallelization}
\label{sec:graphparallelization}

\begin{figure}[t!]
\centering
\includegraphics[width=3.33in]{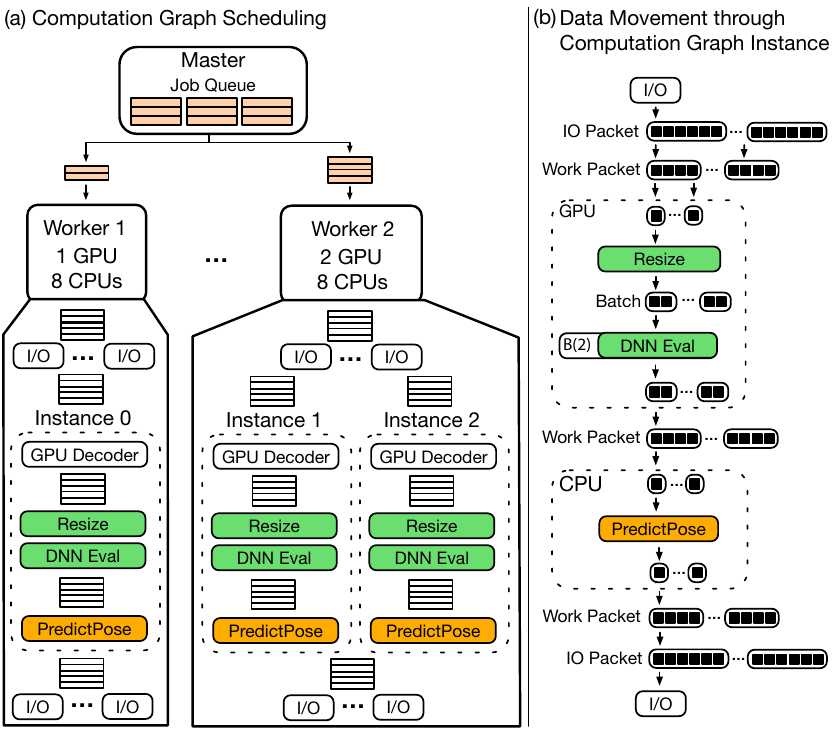}
  \vspace{-1.00em}
\caption{Left: \scanner\ creates multiple computation graph instances to process sequence elements in parallel. Here, three instances of the pose estimation graph (from Fig.~\ref{fig:appsgraphs}-f) are distributed to single-GPU (left) and dual-GPU (right) machines. Instances of I/O and video decode stages that deliver data to and from application-defined graphs are shown in gray. Right: \scanner\ streams data through an execution graph at different bulk granularities to maximize data movement throughput and keep memory footprint low.}
\label{fig:datahierarchy}
  \vspace{-1.5em}
\end{figure}

The \scanner\ scheduler is responsible for efficiently distributing \scanner\ jobs onto the parallel processing resources within a machine and across large clusters of machines. \scanner\ implements data-parallel execution in the presence of stateful kernels by spawning multiple \emph{instances} of the computation graph. In each instance, bounded state graph operations can maintain mutable state buffers, and all graph operations can preallocate a unique copy of read-only buffers (e.g., DNN weights, lookup tables). \scanner\ determines the maximum number of instances that can be created per machine by querying graph operations for their resource requirements, then maximizes parallelism without oversubscribing the machine. Fig.~\ref{fig:datahierarchy}-left depicts a heterogeneous cluster of two machines, each containing an eight-core CPU and at least one GPU (worker 1 contains a single GPU, and worker 2 has two GPUs).  To map the three-stage pose estimation pipeline (Fig.~\ref{fig:appsgraphs}-f), which contains graph operations that require GPU execution and one operation that requires four CPU cores, onto this cluster, \scanner\ creates one computation graph instance on worker 1 and two instances of the pipeline on worker 2.  
 
\scanner\ computation graphs can be statically analyzed to determine each sequence element's dependencies before graph execution. This allows \scanner\ to partition the elements of a job's output sequence into smaller \emph{work packets} without violating graph operation dependencies. Work packets are then distributed to computation graph instances, enabling parallelization within a single video and better load balancing (evaluated in Section~\ref{sec:scalability}). In addition to parallel work distribution, the \scanner\ runtime provides fault tolerance by automatically reassigning and restarting individual work packets (not entire jobs) assigned to failed workers. \scanner\ also distributes work to new worker machines that are added to a cluster while a job is running (supporting elasticity).

\scanner\ implements many common throughput-computing optimizations to sustain high-performance graph execution on machines with many cores and multiple GPUs. These include bulk transfer of sequence data between the data store and video decoders (particularly important in high latency cloud storage scenarios), bulk-granularity time-multiplexing of graph operations onto available machine compute resources, pipelining of CPU-GPU data transfers and data store I/O with graph operation execution (Fig.~\ref{fig:datahierarchy}-right), and using custom GPU memory pools to reduce driver entry-point contention in multi-GPU environments. 

In addition to processing work packets in parallel using multiple graph instances (data parallelism), \scanner\ also parallelizes computation within each graph instance by executing operations simultaneously on different CPU cores and GPU devices (pipeline parallelism). \scanner's current implementation does not distribute the execution of a single graph instance across different machines. (We have not yet encountered applications that benefit from this functionality.) Multi-field elements are provided to operations in struct-of-arrays format to enable SIMD processing by batched operations without additional data shuffling.  The granularities of bulk I/O (I/O packet size) and parallel work distribution (work packet size) are system parameters that can be tuned manually by a \scanner\ application developer to maximize performance, although auto-tuning solutions are possible. We evaluate the benefit of each of these key runtime optimizations in Section~\ref{sec:multigpu}.

\subsection{Unneeded Element Elimination}
\label{sec:depanalysis}

\scanner's sequences are logically dense, however when a computation graph contains sampling operations, only a sparse set of intermediate sequence elements must be computed to generate a job's required outputs.  Since dependencies during graph execution do not depend on the values of sequence elements, \scanner\ determines which elements are required upfront through per-element graph dependency analysis.  Interval analysis methods used to analyze stencil dependencies in image processing systems\,\cite{Ragan-Kelley:2013:Halide} are of little value when required graph outputs span the entire output domain, but are sparse (for example generating every $N$-th frame of an output sequence yields interval bounds that span the entire domain of all upstream sequences).  Instead, given the set of output sequence points a job must produce, \scanner\ analyzes computation graph dependencies to determine the \emph{exact set} of required points for all graph sequences. During graph execution, \scanner\ sparsely computes only the necessary sequence points. During dependency analysis, a bounded state operation with warmup size $W$ is treated like a stencil operation with the footprint ($i-W$,...,$i-1$,$i$).

Fig.~\ref{fig:sampling} illustrates the results of per-element dependency analysis for various example computation graphs. Gray boxes indicate sequence elements that are not required to compute the requested computation graph output elements and do not need to be computed by \scanner.  Performing per-element dependency analysis to identify and eliminate unnecessary computation is unusual in a throughput-oriented system. However, \scanner\ graph operations typically involve expensive processing at the scale of entire frames, so the overhead of computing exact per-element liveness is negligible compared to the cost of invoking graph operations on elements that are not needed for the final job result.

To avoid the storage overhead of fully materializing lists of required sequence domain points, \scanner\ performs dependency analysis incrementally (at work packet granularity) as graph computation proceeds. \scanner\ also coalesces input sequence elements into dense batches to retain the efficiency of batch processing even when dependency analysis yields execution that is sparse.

For all stateless graph operations, sparse execution is a system implementation detail that does not influence the output of a \scanner\ application. It is valid, but inefficient, for \scanner\ to generate all sequence elements, even if they are never consumed. However, since prior invocations of a bounded state operation may impact future output,  the values output by a bounded state operation may depend on which elements the \scanner\ runtime chooses to produce. (Different work distributions or conservative dependency analysis could yield different operation output.)  However, \scanner\ applications are robust to this behavior since bounded state operations by definition are required to produce ``acceptable'' output provided their warmup condition is met.

\subsection{Accessing Compressed Video Frames}
\label{sec:sparsedecode}

\begin{figure}[t!]
\centering
\includegraphics[width=3.333in]{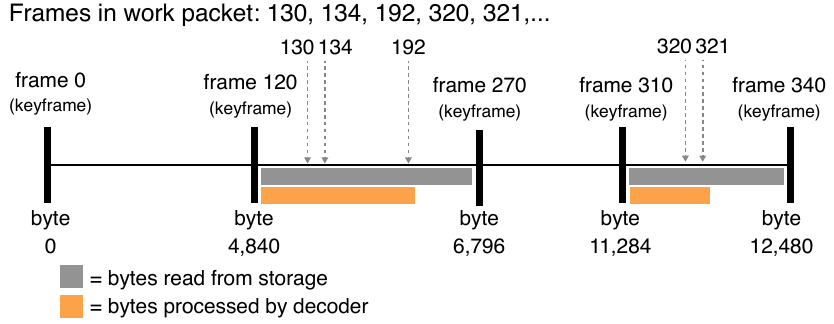}
  \vspace{-1.00em}
\caption{The \scanner\ data store maintains an index of keyframe locations for video frame columns. The index is used to reduce I/O and video decode work when accessing a sparse set of frames from the video.}
\label{fig:decoding}
\vspace{-2.00em}
\end{figure}

\scanner\ presents the abstraction that videos are tables of individual frames, but internally stores video frame columns as compressed H.264 byte streams\,\cite{Marpe:2006:h264} to minimize footprint and to reduce I/O. For example, the footprint of the 12~billion frame \tvnews\ dataset (used in Section~\ref{sec:analyticsapps}) is 20 TB when stored as H.264 byte streams, but exceeds 6~PB when expanded to an uncompressed \mbox{N-D} array of 24-bit pixels.

The cost of supporting compressed video storage in a system that must also support sparse frame-level data access is two-fold. First the byte stream must be decoded on the fly prior to graph execution. Second, video decode involves inherently sequential computation since most frames are encoded as deltas on data in prior frames. Therefore, to materialize a requested video frame, a decoder must locate the preceding ``keyframe'' (the last self-contained frame in the bytestream) then decode all frames up to the requested frame.

To accelerate access and decode of individual frames, the \scanner\ data store maintains an index of the byte stream offsets of keyframes in video columns, similar to indices maintained by video container formats to support scrubbing\,\cite{ISO14496}. The data store uses this index to minimize the amount of I/O and decode performed when servicing a sparse set of frame requests. For example, consider the sequence of elements in Fig.~\ref{fig:decoding}.  To process this sequence, \scanner\ loads bytes from storage beginning from the keyframe preceding frame~130 (at byte offset 4,840).  Decoding begins at this point, and continues until frame~192.  Then, decoder state is reset to keyframe 310, and the process continues. When frames must be decoded but are not required by graph execution (e.g., frames 131-133, 135-191), \scanner\ skips decoder post-processing (extracting frames from the decoder, performing format conversion, etc.).

\scanner's data store implements a number of additional optimizations to maximize throughput, such as avoiding unnecessary reset of video decoder state when multiple required frames fall between two keyframes and time multiplexing decoders at bulk granularity to avoid unnecessary state resets when jobs draw video data from multiple tables. When available, \scanner\ also leverages ASIC hardware decode capabilities to accelerate video decode. For example, use of GPU-resident video decoding hardware frees programmable resources to execute other graph operations and also allows compressed video data to be communicated over the CPU-GPU bus.

\section{Evaluation}
\label{sec:eval}

The goal of \scanner\ is to create a system that is sufficiently expressive to enable a rich set of video processing applications while also maintaining high performance.  We evaluated \scanner's performance in terms of the efficiency of video frame access, efficiency in scheduling computation graphs onto a single machine, and scalability of applications to large numbers of CPUs and GPUs and very large video datasets.  We evaluated \scanner's utility and expressiveness by implementing the video analysis workloads from Section~\ref{sec:workloads} and deploying them at scale.

\subsection{Performance}
\subsubsection{Video Decode Throughput}
\label{sec:decodeperf}
One of \scanner's goals is to provide applications with high-throughput access to compressed video frames, even when requested access patterns are sparse.
We evaluated \scanner's H.264 decode performance against an OpenCV baseline under a varying set of frame access patterns drawn from our workloads:

\begin{itemize}
\item \decode. All video frames

\item \stride. Every 24th frame.

\item \gath. A random list of frames that sparsely samples the video ($0.25\%$ of the video).

\item \range. Blocks of 2,000 consecutive frames, each spread out by 20,000 frames.

\item \keyframe. Only the keyframes from the video.
\end{itemize}

\begin{figure}[tp!]
\centering
\includegraphics[width=3.333in]{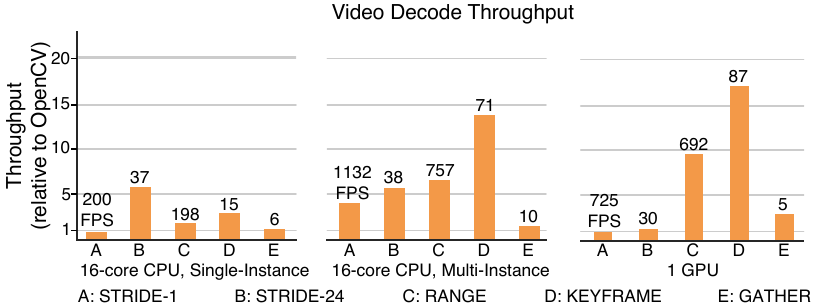}
  \vspace{-1.00em}
\caption{When executing a single graph instance, \scanner's sparse video decode optimizations improve throughput compared to OpenCV baselines on both the CPU and GPU. \scanner\ further improves CPU decode throughput by using multiple graph instances to more efficiently utilize all CPU cores.}
\label{fig:perfdecode}
  \vspace{-1.00em}
\end{figure}

Figure~\ref{fig:perfdecode} presents \scanner's decode throughput under these access patterns on a 2.2 hour, 202,525 frame, 1920$\times$1080 H.264 encoded video (average keyframe distance of 104 frames) on a machine with two 8-core Intel Xeon E5-2620-v4 CPUs and one NVIDIA Titan~Xp GPU. The throughput is normalized to a baseline implementation which makes use of the OpenCV C++ API for video decode on the CPU and GPU (absolute throughput numbers in FPS are also given). The CPU version of this baseline delivers single-machine throughput that is similar to prior work on systems for large-scale video processing\,\cite{Yang:2015:videospark}. For the CPU and GPU, we include results for a single graph instance to isolate the effect of sparse video decode optimizations. For the CPU, we also evaluate multiple graph instances to exploit \scanner's ability to decode different parts of the stream in parallel (we evaluate multiple graph instances on multiple GPUs in Section~\ref{sec:multigpu}).  

In all cases, \scanner's throughput matches or exceeds that of the baselines'. For a single graph instance, \scanner\ realizes higher throughput than the baselines when frame access is sparse (as much as 17$\times$ on the GPU). This speedup comes from \scanner\ avoiding post-decode processing of frames which must be decoded but that are not needed for graph execution (Section~\ref{sec:sparsedecode}). \scanner\ uses the machine's 16 CPU cores more efficiently when executing multiple graph instances (Multi-instance on Fig.~\ref{fig:perfdecode}) since multiple instances of the decoder run in parallel (in addition to the parallelization available in H.264 decode which the baseline also exploits).

Even though \scanner's throughput can be higher than that of the CPU and GPU OpenCV baselines' in sparse access scenarios, overall throughput (FPS) of sparse access is fundamentally lower. If an application is flexible in which frames it can sample, such as accessing only a video's keyframes (\keyframe), it is possible to obtain higher throughput compared to other sparse access patterns (\stride\ or \gath), particularly when decoding on the GPU.

\subsubsection{Scheduling Graphs with Optimized Kernels}
\label{sec:microperf}

In conjunction with video frame access, \scanner\ is also responsible for scheduling computation graphs of optimized kernels to machines with CPUs and GPUs. To test this, we chose three highly optimized kernels drawn from the applications in Section~\ref{sec:workloads} and compared their native performance (when invoked from C++ and using OpenCV for video decode as in Section~\ref{sec:decodeperf}) to \scanner\ implementations using a single compute graph instance. These are:

\textbf{\hist.} Compute and store the pixel color histogram for all frames (video decode bound).  Histogram is computed via OpenCV's \break \code{cv::calcHist}/\code{cv::cuda::histEven} routines on the CPU/GPU respectively.

\textbf{\flow.} Compute optical flow for all frames using a 2-frame stencil (OpenCV's CPU and GPU \code{FarnebackOpticalFlow} routines).

\textbf{\dnn.} Downsample and transform an input frame, then evaluate the Inception-v1 image DNN\,\cite{Szegedy:2015:inception} for all frames.  Image transformation is performed in Halide and DNN evaluation is performed using Caffe\,\cite{Jia:2014:caffe}.

\begin{figure}[tp!]
\centering
\includegraphics[width=3.333in]{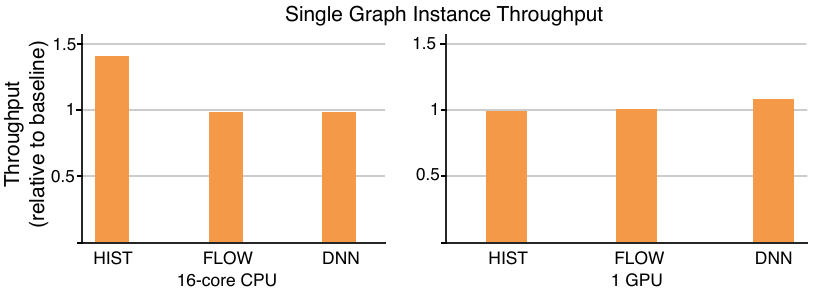}
  \vspace{-1.25em}
\caption{\scanner\ executes graphs implemented using well-optimized kernels with nearly no overhead, matching or exceeding baseline implementations on both the CPU and GPU. Better orchestration of the compute graph produces modest improvements in \hist\ and \dnn.}
\label{fig:perfsinglenode}
  \vspace{-1.00em}
\end{figure}

Figure~\ref{fig:perfsinglenode} presents the throughput of CPU and GPU versions of the \scanner\ implementations of the microbenchmarks (using the libraries given above) normalized to their native implementations. We use the same multi-core CPU + single GPU machine from Section~\ref{sec:decodeperf}. In all cases, the \scanner\ implementations execute the kernels without incurring significant overhead, nearly matching or exceeding the native implementations. The \scanner\ implementations of \hist\ on the CPU and \dnn\ on the GPU achieve a modest improvement in throughput due to better orchestration of the computation graph (pipelining of video decode, data transfers, and kernel execution).




\subsubsection{Single Machine Scalability}
\label{sec:multigpu}


\begin{figure}[tp!]
\centering
\includegraphics[width=3.333in]{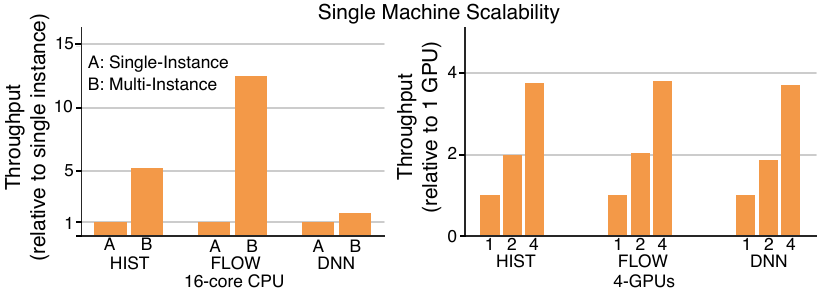}
  \vspace{-1.25em}
\caption{\scanner\ computation graphs can be scaled to machines featuring multi-core CPUs and multiple GPUs without code modification. On the CPU, \scanner\ improves utilization of the 16-cores by parallelizing across multiple frames. On the GPU, \scanner\ achieves near linear speedup (at least 3.7$\times$) when moving from one to four GPUs..}
\label{fig:perfmultigpu}
  \vspace{-1.00em}
\end{figure}

It is common for high-end workstations and modern servers to be packed densely with multiple GPUs and CPUs. We evaluated \scanner's scalability on multi-core CPU and multi-GPU platforms by running the microbenchmarks from Section~\ref{sec:microperf} on a server with the same CPU but now with four Titan~Xp GPUs. Figure~\ref{fig:perfmultigpu} compares the microbenchmarks using \emph{multiple graph instances} against their single graph instance counterparts from Section~\ref{sec:microperf}. Since OpenCV's \hist\ and \flow\ are not parallelized on the CPU, \scanner\ benefits from parallelization across video frames, providing a 5.1 and 12.5$\times$ speedup respectively. Although the Caffe library is internally parallelized, \scanner\ still benefits from processing multiple frames simultaneously for \dnn.

\begin{figure}[tp!]
\centering
\includegraphics[width=3.333in]{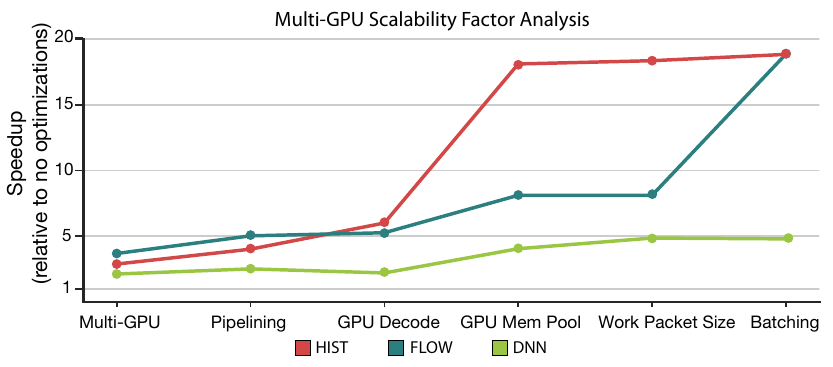}
  \vspace{-1.00em}
\caption{\scanner's runtime optimizations result in a 5 to 19$\times$ speedup of three microbenchmarks on a four-GPU machine. Each microbenchmark benefits differently from the optimizations, but the combination of all optimizations produces the best performance.}
\label{fig:factoranalysis}
  \vspace{-1.00em}
\end{figure}

\begin{figure}[tp!]
\centering
\includegraphics[width=3.333in]{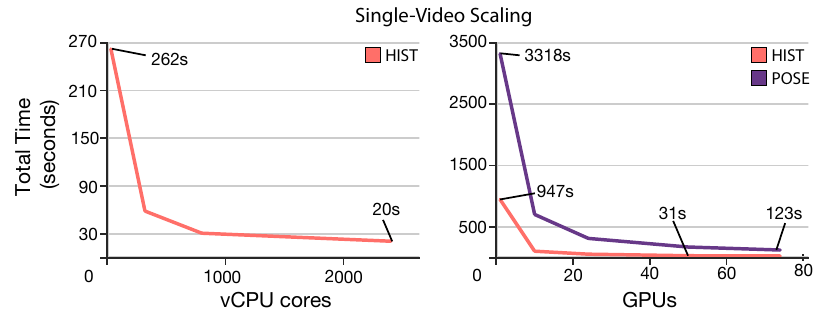}
  \vspace{-1.00em}
\caption{\scanner\ reduces the latency of analyzing a single video by using hundreds of GPUs and thousands of CPU cores.  Scaling out reduces processing times from multiple minutes to seconds.}
\label{fig:interactivescaling}
  \vspace{-1.00em}
\end{figure}

The GPU benchmarks realize near linear scaling (at least 3.7$\times$) from one to four GPUs. The \scanner\ benchmarks realize these throughput improvements \emph{without requiring modification to the \scanner\ application}. Achieving good multi-GPU scaling required the runtime optimizations discussed in Section~\ref{sec:impl}. Figure~\ref{fig:factoranalysis} depicts a factor analysis of these optimizations for the three pipelines used in the four GPU scalability evaluation. The baseline configuration is \scanner\ with all optimizations disabled. Each data point adds one of the optimizations mentioned in Section~\ref{sec:impl}:
\begin{enumerate}
\item Using multiple GPUs
\item Pipelining CPU-GPU computations and data-transfer
\item GPU HW ASIC decode
\item GPU memory pool
\item Increased work packet size
\item Batching input elements to kernels
\end{enumerate}

Even when executing the simple computation graphs of \hist, \flow, and \dnn\ benchmarks, achieving multi-GPU scalability required combining several key optimizations. For example, \hist\ is decode bound, and benefits most from \textbf{GPU Memory Pool} because eliminating per-video frame memory allocations enables the GPU hardware video decoders (enabled by \textbf{GPU Decode}) to operate at high throughput. 
In the case of \dnn, speedups from \textbf{Batching} are only possible after enabling a \textbf{Work Packet Size} that is greater than the batch size.

\begin{figure}[tp!]
\centering
\includegraphics[width=3.333in]{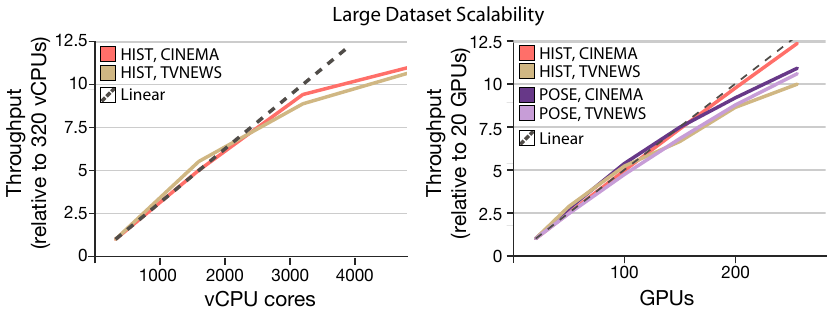}
  \vspace{-1.00em}
\caption{\scanner\ applications efficiently scale to hundreds of GPUs and thousands of CPU cores when processing large datasets. Speedup is nearly linear until stragglers cause reduced scaling at high machine counts.}
\label{fig:batchscaling}
  \vspace{-1.00em}
\end{figure}

\begin{figure*}[tp!]
\centering
\includegraphics[width=7in]{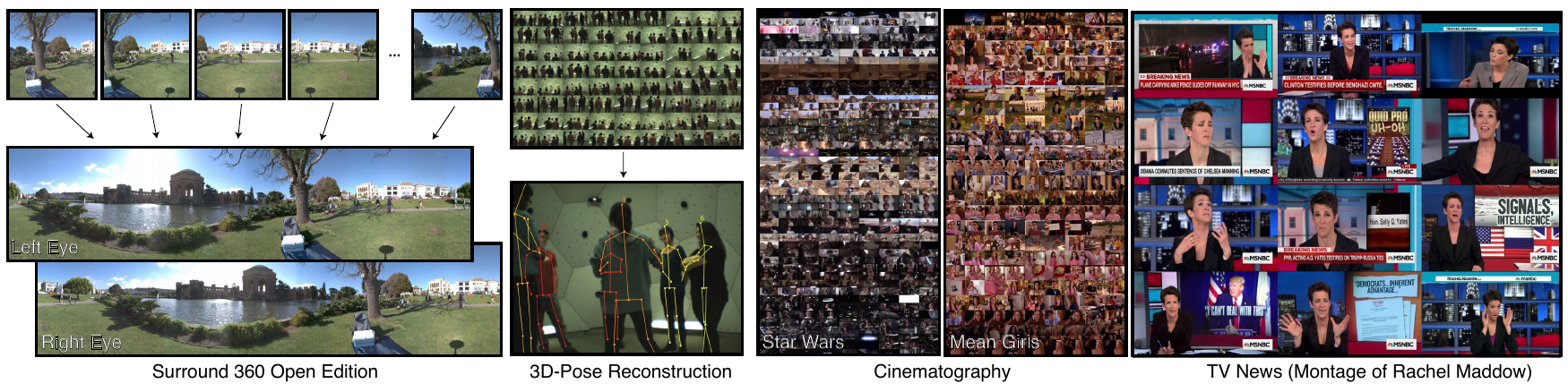}
  \vspace{-1.00em}
\caption{Surround~360: \scanner's port of Surround~360 fuses 14 video streams into a panoramic video for VR display. 3D Pose: Views of a social scene by 72 of the 480 cameras in the CMU Panoptic Studio (Joo et al.\,\protect\shortcite{Joo:2016:Panoptic}). \scanner\ performs pose estimation on all 480 camera streams which are then fused into 3D poses (shown projected onto a single view) Cinematography: A montage of one frame from each shot in \textit{Star Wars} and \textit{Mean Girls} computed using \scanner\ pipelines (Figure~\ref{fig:appsgraphs}-a and -b). TV News: \scanner\ was used to calculate screen time given to people in 70,000 hours of TV News. Here we show instances of Rachel Maddow, a popular news host. Image credit, left to right: "Palace of Fine Arts Take 1", \textcopyright~Facebook 2017; top image from \cite{Joo:2016:Panoptic} Figure 1, \textcopyright~Hanbyul Joo; \textit{Star Wars: Episode IV - A New Hope}, \textcopyright~Lucasfilm Ltd. 1977; \textit{Mean Girls}, \textcopyright~Paramount Pictures 2004; \textit{The Rachel Maddow Show} \textcopyright~MSNBC 2015-2017.}
\label{fig:appsviz}
  \vspace{-1.00em}
\end{figure*}

\subsubsection{Multi-Machine and Dataset Scalability}
\label{sec:scalability}

The true benefit of \scanner\ is the ability to scale video processing applications to large numbers of machines and to very large video datasets. To evaluate \scanner's scalability, we executed two benchmarks, the \hist\ computation graph from Section~\ref{sec:microperf}, and \pose, the OpenPose human pose estimation benchmark~\cite{cao2016realtime} which is central to several larger applications in Section~\ref{sec:applications}, at scale on Google Compute Engine (GCE). We perform CPU scaling experiments on instances with 32 vCPUs (the unit of CPU hardware allocation on GCE, usually one hyper-thread), and GPU scaling experiments on instances with  16 vCPUs and two NVIDIA K80 GPUs. Since the \pose\ benchmark does not support CPU execution, we only evaluate it in GPU scaling experiments.

\textbf{Single Video Scaling.} One use of scaling to large machines is to deliver video processing results back to the user rapidly.  (e.g., for quick preview or analysis.). Figure~\ref{fig:interactivescaling} shows \scanner\ executing \hist\ and \pose\ on a single 2.2 hour feature-length film on a cluster of 2,400 cores and a cluster of 75 GPUs. Executing \hist\ on this video took 4.3 minutes on a single machine (32 vCPUs) and nearly 15 minutes on a single GPU.  These times were reduced to 20 and 31~seconds respectively when parallelizing this computation to large CPU and GPU clusters.  Scaling \pose\ to the large GPU cluster reduced pose estimation processing time from 55~minutes (1 GPU) to two minutes (75 GPUs).

\textbf{Large Dataset Scalability.}
\scanner\ facilitates scaling to large video datasets that would be impractical to process without the use of large numbers of machines. Figure~\ref{fig:batchscaling} shows the speedup achieved running the \hist\ and \pose\ benchmarks on datasets used by the video data mining applications in Section~\ref{sec:analyticsapps}: \cinema, a collection of 657 feature length films (107 million frames, 2.3~TB), and \tvnews, a collection of short clips (approximately 10 seconds each) from 60K TV news videos (these shots total 86 million frames). \scanner\ scales linearly up to 3000 vCPUs and 150 GPUs while continuing to scale near linearly up to 250 GPUs. Speedups are sublinear at higher machine counts since a single slow machine (straggler) can delay job completion. Techniques for mitigating the effect of stragglers are well-studied and could be implemented by a future version of \scanner\ \cite{Ananth:2013:straggler}.

\subsection{Application Experiences}
\label{sec:applications}

We have used \scanner\ to scale a range of video processing applications (Section~\ref{sec:workloads}), enabling us to use many machines to obtain results faster, and to scale computations to much larger video datasets than previously practical. Each application presented a unique combination of frame access patterns, usage of \scanner\ computation graph features, and computational demands.

\newcommand{\todo}[1]{\noindent{\color{red} \textbf{#1}}}

\subsubsection{Video-Based 3D Pose Reconstruction}
\label{sec:poseReconstruction}
The video-based 3D pose reconstruction algorithm by Joo et al.\,\shortcite{Joo:2016:Panoptic} requires efficient scheduling of compute graphs with both CPU and GPU operations to fully utilize machines packed densely with GPUs. The algorithm involves evaluating a DNN on \emph{every frame} of the 480 video streams in the Panoptic Studio (Figure~\ref{fig:appsgraphs}-f). (Per-frame results from each video are then fused to estimate a per-frame 3D pose as in Figure~\ref{fig:appsviz}, 3D Pose). An optimized implementation of the per-frame algorithm took 16.1 hours to process a 40-second sequence of captured video on a single Titan~Xp GPU (frames 13,500 to 14,500 of the ``160422\_mafia2'' scene from the CMU Panoptic Dataset). A version of this algorithm was previously parallelized onto four Titan~Xp GPUs, reducing processing time to seven hours~\cite{cao2016realtime}. Using the exact same kernels, the \scanner\ implementation reduces runtime on the same 4-GPU machine to 2.6 hours due to more efficient graph scheduling (better pipelining and data transfer optimizations as discussed in Section~\ref{sec:multigpu}).

Using \scanner, it was also simple to further accelerate the application using a large cluster of multi-GPU machines in the cloud. The same \scanner\ application scheduled onto 200 K80 GPUs \linebreak (25 8-GPU machines on GCE) completed processing of the same video sequence in only 25 minutes. Dramatically reducing pose reconstruction time to minutes stands to enable researchers to capture longer and richer social interactions using emerging video-based capture infrastructure such as the Panoptic Studio.


\subsubsection{Hyperlapse Generation}
\label{sec:hyperlapse}
The real-time hyperlapse algorithm of \cite{Joshi:2015:hyperlapse}, which computes stabilized timelapses, makes use of computations that stencil over temporal windows. The computational bottleneck in the hyperlapse algorithm is feature extraction from the input images and pairwise feature matching between neighboring images. We implemented those portions of the algorithm as kernels in \scanner\ (Figure~\ref{fig:appsgraphs}-d) using a GPU kernel to extract SIFT features from each frame and a second GPU kernel with a stencil window of size $w$ to perform feature matching. \scanner's stenciling mechanism simplified the implementation of the feature matching kernel (the runtime handles storing intermediate video frames and results) and made the pipeline easy to extend. For example, Joshi et al.~\shortcite{Joshi:2015:hyperlapse} suggest a performance optimization that approximates the reconstruction cost between two frames as the sum of successive costs, falling back to the full windowed feature matching when necessary. The corresponding \scanner\ pipeline (Figure~\ref{fig:appsgraphs}-e) reduces the matching kernel's stencil size to $[0,1]$ to capture the adjacent reconstruction costs and adds a new kernel \verb|CondMatch| which stencils over both the derived matching costs and the original features, conditionally determining if it is necessary to perform the full windowed feature matching.


\subsubsection{Visual Data Mining at Scale}
\label{sec:analyticsapps}
We have also used \scanner\ as the compute engine for two big video data mining research projects, requiring sparse sampling of videos, bounded state, and fault tolerance when scaling to hundreds of machines. The first involves visual analysis of a corpus of 657 feature length films (7.7 million frames, 2.3~TB). For example, \scanner\ applications are used to detect shot boundaries (via histogram differences, Figure~\ref{fig:appsgraphs}-a), produce film summaries via montage (as in Figure~\ref{fig:appsviz}-middle, with the \scanner\ pipeline in Figure~\ref{fig:appsgraphs}-b), and detect faces.  The second is a large-scale analysis of video from over three years of US TV news (FOX, MSNBC, and CNN), which includes over 70,000 hours of video (20~TB, 12~billion frames, six petapixels). In this project \scanner\ is being used to perform large-scale data mining tasks to discover trends in media bias and culture. These tasks involve visual analyses on video frames such as classifying news into shots, identifying the gender and identity of persons on screen, estimating screen time of various individuals, and understanding the movement of anchors on screen via pose estimation. Use of \scanner\ to manage and process billions of video frames was essential.

The large size of the feature length films and the TV news dataset stress-tested \scanner's ability to scale. For example, to estimate the screen time allotted to male-presenting versus female-presenting individuals, we used \scanner\ to compute color histograms on every frame of the dataset (to detect shot boundaries), and then sparsely computed face bounding boxes and embeddings on a single frame per shot. To execute these tasks, we used a GCE cluster of 100 64-vCPU preemptible machines, relying on \scanner's fault tolerance mechanism to handle preemption. The size of the dataset also required the use of cloud storage for both the videos and the derived metadata. Each computation took less than a day to complete and \scanner\ maintained 90\%+ utilization of the 6,400 vCPUs throughout each run.

\subsubsection{VR Video Stitching}
\begin{figure}[tp!]
\centering
\includegraphics[width=3.333in]{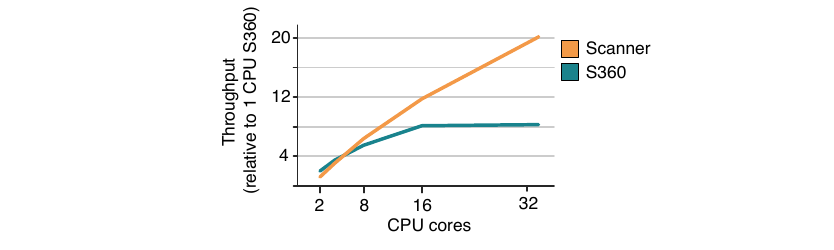}
  \vspace{-1.25em}
\caption{Single-machine scalability of the Surround~360 pipeline implemented in \scanner\ vs. the open source implementation. The \scanner\ implementation utilizes a single-machine well, scaling better to higher core counts.}
\label{fig:s360pipeline}
  \vspace{-1.50em}
\end{figure}


We ported the Facebook Surround~360 Open Edition VR video stitching pipeline to \scanner\,\cite{Surround360}. The application requires simultaneously accessing 14 input video streams, scheduling up to 44 computation graph operations on a large number of CPU cores, employing kernels with temporal dependencies (the \emph{Flow} kernel is configured as a bounded state operation since it depends on the output of previous frames), and compressing output video frames to produce the final stereo panorama output (Figure~\ref{fig:appsgraphs}-c). Given \scanner's current scheduler implementation, we found it most efficient to execute each \emph{Warp}, \emph{Flow}, \emph{Synth} block (the kernels surrounded by the blue box in Figure~\ref{fig:appsgraphs}-c) as a separate job in \scanner\ and then feed each of those job's outputs into the \emph{Concat} stages using a second bulk launch. The \scanner\ implementation uses the same kernels as Facebook's reference implementation.

In contrast to the reference Surround~360 implementation, which is parallelized across the 14 input video streams (but outputs frames serially), our \scanner\ implementation is also parallelized across segments of output frames, making use of bounded state operations with warmup of size 10 to maintain temporal coherence across segments of the video. Figure~\ref{fig:s360pipeline} plots the relative speedup of the reference and \scanner\ Surround~360 implementations on a machine with 32 CPUs (64 hyper-threaded). The \scanner\ implementation scales more efficiently on the large machine (5.3 seconds per frame versus 13.3 seconds per frame for the reference) due to the change in parallelization strategy. It is also faster due to pipelining (overlapping data movement and compute) and decreased IO since the \scanner\ implementation performs compression of the large output frames on the fly before writing out to disk.

We ran the \scanner\ version of Surround~360 implementation on a one minute sequence (28 GB, 25k total frames) over eight machines with 32 vCPU cores each (256 cores total) on Google Compute Engine and achieved a rate of 1.5 FPS. As was the case with our other applications, we were able to scale Surround~360 without any changes to the \scanner\ application.


\section{Related Work}
\label{sec:background}

\scanner\ contributes a unique integration of data-flow
  programming abstractions and systems implementation components that
  meet the productivity and performance needs of video analysis
  applications.  However, many \emph{individual components} of \scanner's design
  were influenced by prior systems for big data processing, databases,
  and machine learning.

\textbf{Distributed data analytics frameworks.}  Frameworks such as
MapReduce\,\cite{Dean:2004:MapReduce} and
Spark\,\cite{Zaharia:2010:Spark} enable concise and productive
expression of data analytics applications using data parallel
operations on large collections. While these platforms handle the
``scale-out'' scheduling challenges of distributed computing (e.g. work distribution and fault tolerance), as identified in Section~\ref{sec:challenges}, they require new primitives and significant changes to their internal implementation to meet a broad set of video analysis needs.
For example, while it is possible to use Spark to process video, prior implementations\,\cite{Yang:2015:videospark} do not implement intra-video parallelism (precluding single-video speedups), do not target heterogeneous machines, and do not implement the video decode optimizations shown to provide significant benefits in Section~\ref{sec:decodeperf}.  \scanner\ features such as bounded state operations (needed for intra-video parallelization in applications like VR video synthesis) and unneeded element elimination (needed for efficient sparse sampling common in data mining, Sec.~\ref{sec:analyticsapps}) do not yet exist in popular distributed data-parallel systems.

Also, as we demonstrate in Fig.~\ref{fig:factoranalysis}, \scanner\ execution graphs require a high-performance, heterogeneous (CPU, GPU, ASIC) runtime to be executed efficiently. While recent efforts have exposed popular GPU-accelerated machine
learning libraries\,\cite{TensorFrames,CaffeOnSpark} to Spark
applications, the Spark runtime, including its task scheduling,
resource management, and data partitioning decisions, operates with no
knowledge of the heterogeneous capabilities of the platform.
Extending Spark to schedule tasks onto high-throughput accelerated
computing platforms is known to require significant runtime redesign
and extensions to application-visible abstractions (e.g., ability for
kernels to specify resource requirements and data layouts, and to
maintain local state)\,\cite{Tungsten,Bordawekar:2016:sparkgpu}.
We hope that the design and implementation of
\scanner\ influences ongoing development of the Spark runtime to
better support video processing applications and accelerated computing.

\textbf{Distributed machine learning frameworks.}  Modern machine
learning frameworks\,\cite{TensorFlow:2016,MxNet:2015,MicrosoftCNTK}
adopt a general dataflow programming model suitable for distributing
GPU-accelerated training and inference pipelines across clusters of
machines.  While it may be possible to implement \scanner's
functionality as a library built upon these frameworks, doing so would
require implementing new operations, runtime support for media
accelerators, and integration with a pixel storage system providing
the desired relational model and efficient video access---in other
words, reimplementing most of \scanner\ itself. We elected to
implement \scanner\ from the ground up as a lightweight runtime for
simplicity and to achieve high performance.

\textbf{Databases for raster and array data.} \scanner\ models image
and video collections as relations, and echoes the design of
SparkSQL\,\cite{Armbrust:2015:SparkSQL} in that row selection and joins
on relations are used to define distributed datasets streamed
to processing pipelines.  Like relational Geographic
Information Systems (GIS) (\cite{PostGISManual}) or
science/engineering-oriented Array Databases (ADBMS) such as
SciDB\,\cite{Cudre-Mauroux:2009:SciDB} or
RasDaMan\,\cite{Baumann:1998:RasDaMan} which extend traditional
database management systems with raster or multi-dimensional array
types, \scanner\ natively supports image and video column types for
efficiency.  While GIS and ADBMS are optimized
for operations such as range queries that extract pixel regions from high-resolution images, \scanner's storage
layer is designed to efficiently decompress and sample sequences of frames for delivery to computation graphs.
As stated in Section~\ref{sec:challenges}, in contrast to array database designs, we intentionally avoided creating a new language for processing pixels in-database (e.g., SciDB's Array Functional Language or RasDaMan's RASCAL\,\cite{RasDaMan:Rascal}). Instead we chose to support efficient delivery of video frame data to execution graphs with operations written in existing, well-understood languages like CUDA, C++, or Halide\,\cite{Ragan-Kelley:2012:Halide}.

\section{Discussion}
\label{sec:discussion}

As large video collections become increasingly pervasive, and algorithms for interpreting their contents improve in capability, there will be an increasing number of applications that require efficient video analysis at scale.  We view \scanner\ as an initial step towards establishing efficient parallel computing infrastructure to support these emerging applications.  Future work should address higher-level challenges such as the design of query languages for visual data mining (what is SQL for video?), the cost of per-frame image analysis for the case of video (e.g., exploiting temporal coherence to accelerate DNN evaluation on a video stream), and integration of large-scale computation, visualization, and human effort to more rapidly label and annotate large video datasets\,\cite{Ratner:2018:Snorkel}.

While the current version of \scanner\ achieves high efficiency, it requires the application developer to choose target compute platforms (CPU vs. GPU), video storage data formats, and key scheduling granularities (e.g., task size).  It would be interesting to consider the extent to which these decisions could be made automatically for the developer as an application runs.   Also, simple extensions of \scanner\ could expand system scope to provide high-throughput delivery of sampled video frames in model training scenarios (not just model inference) and to deliver regions of video frames rather than full frames (e.g., to support iteration over scene objects rather than video frames).

Most importantly, we are encouraged that \scanner\ has already proven to be useful.  Our collaborations with video data analysts, film cinematographers, human pose reconstruction experts, and computer vision researchers show \scanner\ has enabled these researchers to iterate on big video datasets much faster than before, or attempt analyses that were simply not feasible given their level of parallel systems experience and existing tools. We hope that \scanner\ will enable many more researchers, scientists, and data analysts to explore new applications based on large-scale video analysis.

\section{Acknowledgments}

This work was supported by the NSF (IIS-1422767, IIS-1539069), the
Intel Science and Technology Center for Visual Cloud Computing, a
Google Faculty Fellowship, and the Brown Institute for Media
Innovation.  TV News datasets were provided by the Internet Archive.

\bibliographystyle{ACM-Reference-Format}
\bibliography{scanner}

\pagebreak
\section{Appendix A}

\subsection{Video Representations}
\label{sec:videorepresentation}

As discussed in Section~3.1, representing videos as tables allows \scanner\ to decouple the \emph{logical} representation of a video (each frame a distinct row in a table) from the \emph{physical} storage format. In this section, we will show how the table representation enables high throughput video decoding and eases management of the video representation by exploring a variety of \emph{physical} video formats that are all accessed using the same \scanner\ table interface. Due to the flexibility of the execution engine, we were able to perform all of the following storage format transformations directly within \scanner.

\begin{figure}[tp!]
\centering
\includegraphics[width=3.333in]{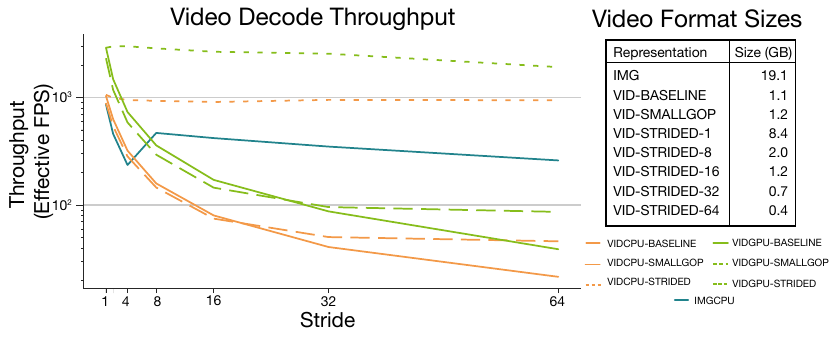}
\caption{Left: Effective throughput of various video representations at increasing stride. The evaluation was run on a machine with two 8-core Intel Xeon E5-2620 CPUs and four Pascal Titan Xp GPUs. Right: A table with the on-disk size of each video representation.}
\label{fig:decodetradeoffs}
\end{figure}

Figure~\ref{fig:decodetradeoffs} shows the throughput in frames per second of decoding frames from different physical video formats of the same videos. The evaluation was run on three 1920x1080 feature length films (a total of 600k frames). The size of each representation is listed in the table of Figure~\ref{fig:decodetradeoffs}. In the following paragraphs, we will walk through the tradeoffs associated with each format under different access patterns.

\textbf{Images.} \imgcpu\ represents reading 95\% quality JPEG images pre-extracted from the video. Images can be read and decoded independently of each other so they provide good performance for sparse access patterns. However, images have a significantly larger storage footprint (170GB vs 1.2 GB for H.264 video) and are thus bound by I/O throughput.

\textbf{Video.} \vidcpu-\decvideo\ and \vidgpu-\decvideo\ show H.264 video decode on the original video format. The low stride performance is high and the storage footprint is low. However, since decoding a specific frame in video can require decoding all preceding frames in a keyframe sequence (tens to hundreds of frames), \scanner\ must decode an increasing percentage of unused frames as the stride increases.


\textbf{Video with shorter keyframe intervals.} Video decode throughput at higher strides can be improved by decreasing the distance between keyframes, trading off an increase in file size (more keyframes consume more storage space). This is shown by the improvement in throughput at large strides and increase in file size of \vidcpu-\decsmallgop\ and \vidgpu-\decsmallgop\ which perform decode on a video table that was re-encoded using \scanner\ with a keyframe interval of 24.

\textbf{Strided Video.}  If an access pattern is known a prior and expected to occur multiple times, higher throughput can be achieved by preprocessing videos to extract and re-encode the specific frames of interest. \vidcpu-\decstrided\ and \vidgpu-\decstrided\ show higher decode throughput and small on disk storage size for video decode over videos preprocessed using a \scanner\ pipeline that selected the desired frames at the given stride and re-encoded them as a new table (since the size for \decstrided\ changes with stride, we listed its size for each stride chosen).
 
The above experiments show there is a continuum of storage formats for video and and that the \scanner\ table abstraction allows a user to explore them easily.


%

\end{document}